# Dynamics Estimation Using Recurrent Neural Network

Astha Sharma

University of South Florida

I. ABSTRACT

There is a plenty of research going on in field of robotics. One of the most important task is dynamic estimation of response during motion. One of the main applications of this research topics is the task of pouring, which is performed daily and is commonly used while cooking. We present an approach to estimate response to a sequence of manipulation actions. We are experimenting with pouring motion and the response is the change of the amount of water in the pouring cup. The pouring motion is represented by rotation angle and the amount of water is represented by its weight. We are using recurrent neural networks for building the neural network model to train on sequences which represents 1307 trails of pouring. The model gives great results on unseen test data which does not too different with training data in terms of dimensions of the cup used for pouring and receiving. The loss obtained with this test data is 4.5920. The model does not give good results on generalization experiments when we provide a test set which has dimensions of the cup very different from those in training data.

II. INTRODUCTION

Dynamic estimation of response is done by humans in all day to day activities like walking, grabbing objects, pouring things, etc. We learn to estimate dynamic responses by experience and this is done subconsciously by us. Researchers are working on training robots for such things to increase their usability. One of the main task for which researches are going on is pouring [1] which is one of the most commonly used tasks in cooking. For doing this task, artificially intelligent model would need information like rotation angle, shape and size of cup and information about the liquid like its density. Such approach of training robots by providing examples is called programming by demonstration (PbD) [2].

There are many researches are going on for proper trajectory optimization. It has a wide range of applications specially in robotics. Estimation of appropriate weight while pouring have been active research area in robotics. Dynamic estimation of parameters plays important role in determination of the parameter.

There are many different trajectory generation methods which includes dynamic estimation of response. One of them is Dynamic Movement Primitives (DMP), which is a non-linear dynamical system which can model discrete motions like swinging tennis racket, walking etc. [4] [1]. Another method is Gaussian Mixture Model (GMM) which is used to generate trajectories of movements [5] [6]. Principal Component Analysis (PCA) is another very useful and very commonly used approach for motion trajectory generation. It utilizes dimension reduction and can predict how a motion changes with time [7]. Another approach is object–object-interaction affordance knowledge to perform object classification, action recognition and manipulation of task. "This method can connect and model the motion and features of an object in the same frame" [8]. The training is done with labeled video sequences and represented as a Bayesian Network. Bayesian Network includes objects, human action, and object reaction as parameters [8].

We used recurrent neural network for our experiments. RNN is being very popular for sequential data. It takes a give input and the output produced by previous step as input and generates an output which is fed to the next step. Recurrent neural network is good when we work on time series. It is suitable for spetio-temporal information processing [3]. That makes RNN good for trajectory estimation. RNN can be applied on wide verity of data such as data involving time sequences as well as ordered sequences like characters in music synthesis and financial forecasting. The model which can be useful for such hard tasks need to have dozens of layers thousands of parameters to train well. Complexity, overfitting and requirement of large dataset are downsides of such very deep networks space.

This report is based on the experiments done to help for the task of dynamic estimation of response and to make a good model in terms of loss and precision while pouring. The experiments were performed on the data set that include 1,307 motion sequences and their corresponding weight measurements. The task was to create a model which takes information of the cup, liquid, pouring angle etc. As input and produces respective weight of liquid which can be poured without spill, as output. We designed a recurrence neural network which estimates weight of the liquid which can be poured into the cup. We used the RNN structure, Long Short-Term Memory used for training. Multiple layers of LSTM, dropout and fully connected layers are used to prepare the model. The model was optimized twice using Adam optimizer.

III. DATA AND PREPROCESSING

Deep learning uses neural networks with hundreds of hidden layers and it requires large amounts of training data. Building a good neural network model always requires careful consideration of the architecture of the network as well as the input data format. Data collection and preprocessing are very crucial parts of deep learning related experiments.

The data provided for the experiment is the set including 1307 motion sequences which represent different 1307 trails. The motion sequences include following information:

$\theta_t$ = rotation angle at time t (degree)
$f_t$ = weight at time t (lbf)
$f_{init}$ = weight before pouring (lbf)
$f_{empty}$ = weight while cup is empty (lbf)
$f_{final}$ = weight after pouring (lbf)
$d_{cup}$ = diameter of the receiving cup (mm)
$h_{cup}$ = height of the receiving cup (mm)
$d_{ctn}$ = diameter of the pouring cup (mm)
$h_{ctn}$ = height of the pouring cup (mm)
☐ = material density / water density (unitless)

Only $\theta_t$ and $f_t$ changes with time, all other values remain unchanged throughout the sequence. Weight at time t ($f_t$) need to be predicted using rotation angle at time t ($\theta_t$) The shape of data is [num sequence, max len, feature dim]. All the sequences have different max len value (maximum length), so the max len is padded with zeros up to 1099 to make all the length equal.

Pre-processing of data is a technique that involves transforming raw data into a managed and consistent format. Data can be incomplete, inconsistent, or lack certain behaviors [16]. Data preprocessing helps to resolve such issues. This technique prepares raw data to be processed further as the input to the neural network [17].

For our experiments we did basic preprocessing by partitioning data into training, validation and testing datasets. A good practice for partitioning data is to shuffle data before partitioning to get an unbiased distribution of data into training and testing sets [18]. For this project, data partitioning is done with the ration 4:1. It means 80% of data is kept for training and remaining 20% is kept for testing. This 20% data is further partitioned as 70% validation data and 30% test data. It means out of 1045 sequences are used for training, 183 for validation and 79 for testing.

IV. METHODOLOGY

The detailed description of model and training approach is following:

A. RNN Architecture

RNN is great for sequential data because each neuron acts as an internal memory to store information of previous input. It performs computation one step at a time sequentially [10]. Figure 1 shows basic structure and working of RNN.

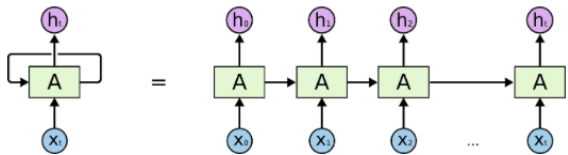

Figure 1: Basic structure of RNN [8]

$$h_t = A(w[x_t + h_{t-1}] + b_t)$$

here, $h_t$ is the output of activation function A at given time step t which takes weight (w) and bias ($b_t$) as input. $x_t$ is the input to the neural network at given time step and $h_{t-1}$ is the output of previous step. The huge depth of RNN helps them to learn efficiently. However, this large depth makes RNN difficult to train because of exploding and vanishing gradient problem [9] [11] [12]. Due to the vanishing gradient problem, it becomes very difficult for RNN to learn and tune parameters of previous layers. Long Short-Term Memory (LSTM) and Gated Recurrent Unit (GRU) are improved versions of RNN which are commonly used and solves the vanishing gradient problem. These networks can remember information for longer period and hence solves the long-term dependency problem. Bothe networks can remove or add information by structures called gates. Gates help to remember or forget information. GRU has two gates (reset and update gates) as compared to LSTM which has three gates (input, output and forget gates) [13]. It can be said that LSTM has have more control on the network because of more number of gates, but GRU is faster to train because it doesn't need memory unit. LSTM is said to be better with large data and it is more stronger and generalized than GRU, but there is no certain guideline about which network should be used in which situation [14] [15]. In our experiments we have used both LSTM and GRU and compared the results. LSTM gave better results for our experiments. Working mechanism of LSTM is shown in figure 2.

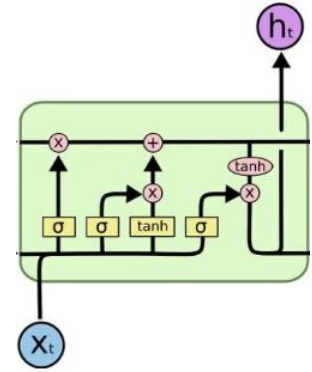

Figure 2: Basic structure of LSTM [24]

$i_t = \sigma (W_i [h_t - 1, x_t]^T + b_t)$
$f_t = \sigma (W_f [h_t - 1, x_t]^T + b_f)$
$o_t = \sigma (W_o [h_t - 1, x_t]^T + bo)$
$\acute{c}_t = \sigma (W_o [h_t - 1, x_t]^T + b_{\acute{c}t})$
$c_t = ft \odot ct-1 + i_t \odot \acute{c}_t$
$h_t = o_t \odot tanh(c_t)$

Here i, o, f and c are the input, output and forget gates and cell respectively at t time step. σ represents sigmoid activation function, ⊙ represents element wise multiplication and $h_t$ is the output of hidden later at time step t.

B. Proposed Model Structure

The model created to be trained on the dataset in this project has 5 LSTM layers, 2 dense layers and 1 dropout layer (Figure 2).

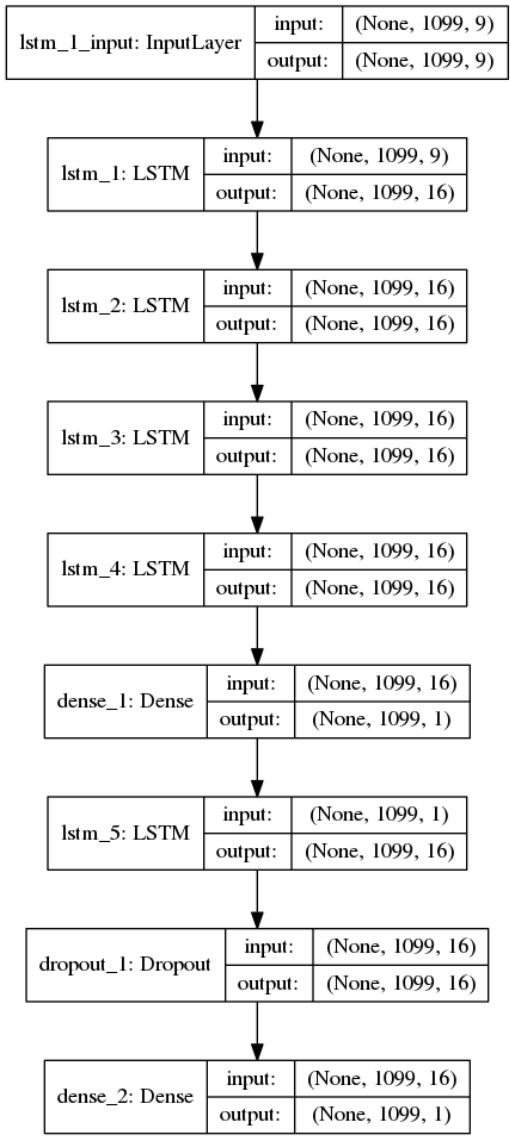

Figure 2: Architecture of the proposed model

The first four layers of the network are LSTM layers. Input shape and dimensions must be determined in the first LSTM layer. LSTM layers has three gates (input, output and forget gates), and present input and past output is fed to each of these three gates which will handle the input [13]. Activation function used with LSTM layers in tanh. The range of tanh function is from (-1 to 1) which gives it advantage over other activation functions because of its larger range. In addition to it, tanh has the property that its second derivative can endure for a long range before becoming zero [19]. This is necessary to overcome vanishing gradient problem. For recurrent activation we used hard-sigmoid. An exponential function is not needed during this step. We used hard-sigmoid for this because it is faster to calculate. The reason for fast speed is that is it is computationally cheaper. return_sequence decides that the last output in the output sequence or the full sequence should be returned. We kept it true to return the full sequence because the LSTM layer expects full sequence as input. Next layer is a fully connected or dense layer, which reduces the output dimension to one, which helps making the network more complicated so that it can become able to learn efficiently. The activation function used in this layer is Rectified Linear Unit (ReLU). ReLU has become very popular in recent years and it is proved to have 6 times improvement in convergence from Tanh function [16]. Mathematical formula of ReLU is:

$$R(x) = max\,(0, x)$$
$$i.e.\ if\ x < 0,\ R(x) = 0$$
$$and\ if\ x >= 0,\ R(x) = x$$

ReLU used as an activation function have been highly successful for computer vision tasks and gives better speed and performance that standard sigmoid function [16].

The next layer is dropout, which is used for regularization. Regularization helps the model to generalize better by reducing the risk of overfitting. There is a risk of overfitting if the size of data set is too small as compared to the number of parameters needed to be learned. A dropout layer randomly removes some nodes and their connections in the network. We can use dropout with hidden or input layer. By using dropout, the nodes in the network become more insensitive to the weights of the other nodes. This makes the model robust [21] [22]. We are keeping the dropout rate 0.2%, that means one of 5 inputs will be randomly ignored during calculation. Keeping dropout rate up to 0.2% is considered optimal performance with small data sets [22].

Last layer in the network is a fully connected layer with ReLU activation function again. Acting as the output layer, the fully connected layer reduces the input dimensions again to one.

### C. Training

The task was to predict weight ($f_t$) at time t using rotation angle ($\theta_t$) at time t. The input sequence has extra zeros padded the end of all features. These zeros are not actual data and need to be omitted during error calculation. Similarly, extra garbage values, which are added to the predicted vector because of the zeros padding, also need to be omitted in error calculation. Experiments were done using Mean Square Error and Euclidean Distance loss functions. Both loss functions gave similar results while testing, but MSE shows faster convergence, which may result to overfitting. So, to prevent overfitting, we chose Euclidean Distance loss for calculating error. Euclidean Distance between vectors $y_{true}$ (actual output) and $y_{pred}$ (predicted output) is calculated by taking the square root of sum of square of distances between all n points in vectors $y_{true}$ and $y_{pred}$, where, n is the length of vectors [25].

$$Euclidean\ Distance = \sqrt{[(y_{true1} - y_{pred1})^2 + (y_{true2} - y_{pred2})^2 + (y_{true3} - y_{pred3})^2 + \ldots + (y_{truen} - y_{predn})^2]}$$
$$Euclidean\ Distance = \sqrt{\sum (y_{true} - y_{pred})^2}$$

Experiments were done using Adam optimizer. Adam results good for data set with multi-dimensional parameters and works well on wide range of problems. It computes adaptive learning rates for all the parameter [27]. It is robust and proved to give better results than many other optimizers like RMSProp, Adadelta and AdaGard, because in addition to storing an exponentially decaying average of past squared gradients, Adam also keeps an exponentially decaying average of past gradients [26] [27]. The learning rate was set to 0.0001 because lower learning rate works better for slow convergence and hence avoids overfitting. The training was done for 2000 epochs. The total number of parameters trained during the process were 9,186.

Summary of the trained neural network model is shown below (Figure 3).

```
Layer (type)                 Output Shape              Param #
=================================================================
lstm_1 (LSTM)                (None, 1099, 16)          1664
lstm_2 (LSTM)                (None, 1099, 16)          2112
lstm_3 (LSTM)                (None, 1099, 16)          2112
lstm_4 (LSTM)                (None, 1099, 16)          2112
dense_1 (Dense)              (None, 1099, 1)           17
lstm_5 (LSTM)                (None, 1099, 16)          1152
dropout_1 (Dropout)          (None, 1099, 16)          0
dense_2 (Dense)              (None, 1099, 1)           17
=================================================================
Total params: 9,186
Trainable params: 9,186
Non-trainable params: 0
```

Figure 3: Summary of the proposed model

### D. Input and Output

Input provided to the network is a 3-dimenssional array of tensors with shape [num sequence, max len, feature dim] representing sequence of rotation angles changing with time step. The output is 2-dimentional array of tensors with shape [num sequence, feature dim] representing sequence of predicted weights for every rotation angle.

## V. EVALUATION AND RESULT

### A. Testing, Evaluation and Results

Various experiments were done during training to produce the best model. Combination of different training options were used which include trying different types of RNN (LSTM and GRU), loss functions (MSE and Euclidean Distance), Activation functions (linear and ReLU). Results obtained with these experiments are shown below (Table 1).

| Training Options Combination | Training Loss | Validation Loss | #epochs |
|---|---|---|---|
| RNN: LSTM Loss: MSE Activation function: Linear | 0.0076 | 0.0068 | 500 |
| RNN: LSTM Loss: Euclidean distance Activation function: ReLU | 14.464 | 12.4351 | 500 |
| RNN: LSTM Loss: MSE Activation function: ReLU | 0.00476 | 0.00394 | 1000 |
| RNN: LSTM Loss: Euclidean distance Activation function: Linear | 6.7094 | 5.4249 | 1500 |
| RNN: GRU Loss: Euclidean distance Activation function: ReLU | 15.8974 | 15.1060 | 500 |
| RNN: GRU Loss: MSE Activation function: ReLU | 0.0087 | 0.0078 | 500 |
| RNN: LSTM Loss: Euclidean distance Activation function: ReLU | 5.9478 | 4.5920 | 2000 |

Table 1: Results generated with various experiments done during training.

From the results in Table 1, it can be deduced that LSTM gives better results than GRU for the provided data set. While comparing linear and ReLU activation functions, ReLU performed slightly better. If we talk about loss functions, Euclidean distance converges slower and gives better results than MSE. Hence the experiment with LSTM was finalized for training with Euclidean distance loss and ReLU activation function. The final training and validation losses obtained with this combination were 5.9478 and 4.5920 respectively.

Testing of the generated model was done twice.

Test 1: With the dataset which we partitioned and kept aside during pre-processing of input data

Test 2: With another dataset provided for testing to test generalization.

Both data sets are unseen to the network.

### I. Test 1

This dataset has 79 trail sequences. Because this test set is partitioned from the input data itself, dimensions of the cup (height and diameter) is not so different in these sequences than dimensions in the training sequences. Graphs to show comparison between actual and predicted values of weight ($f_t$) during some sequences are shown below (Figure 6).

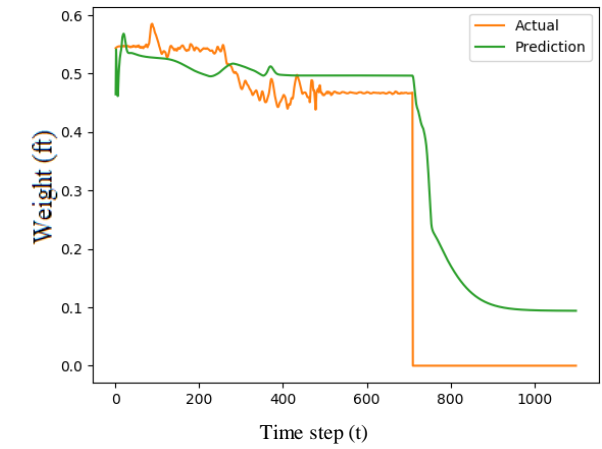
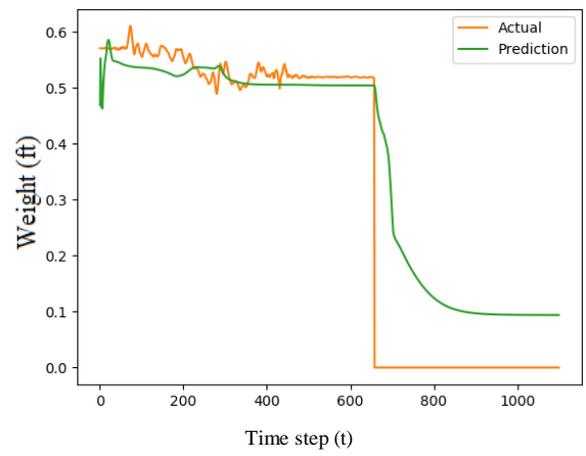
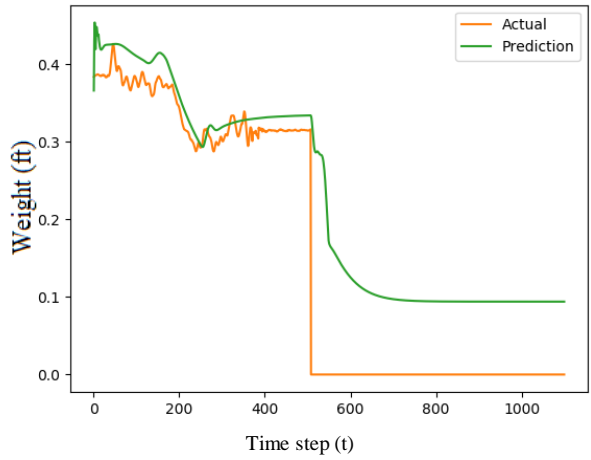
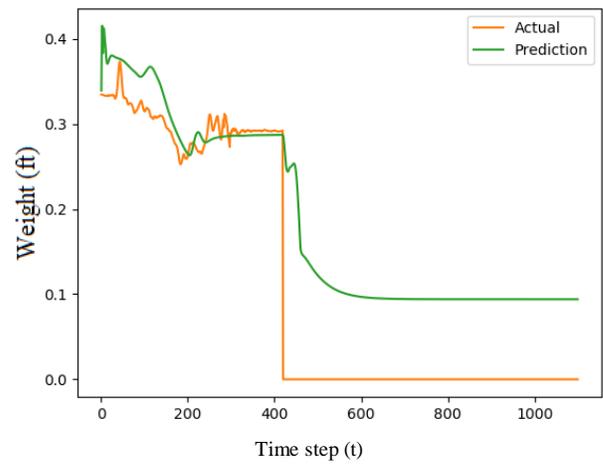
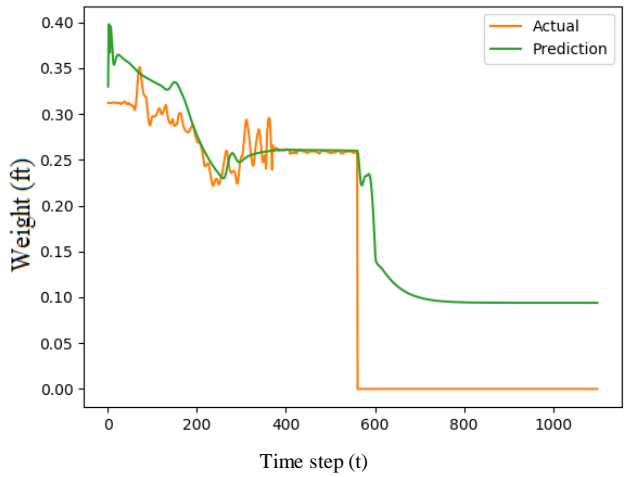
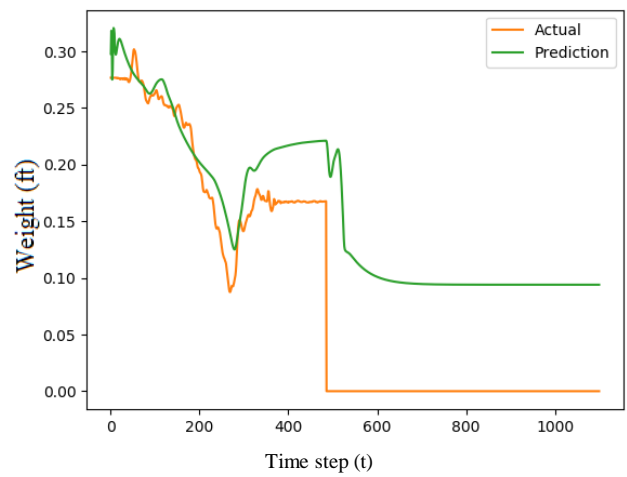

Figure 6: Actual and predicted values of weight ($f_t$) using 79 test sequences kept aside during data partitioning.

## II. Test 2

To test generalization, further testing is done on a different test set with 289 sequences. Dimensions of the cup (height and diameter) is very different in these sequences than dimensions in the training sequences. Results are plotted for sequence number 286, 18, 10, 171, 267 and 203. Graphs to show comparison between actual and predicted values of weight ($f_t$) for respective sequences are shown below (Figure 7).

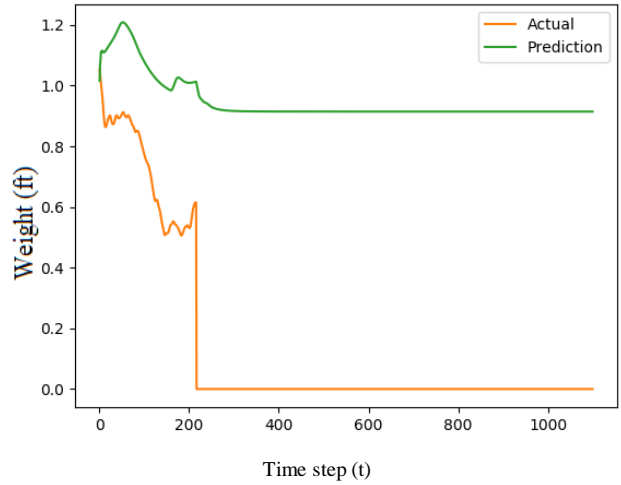
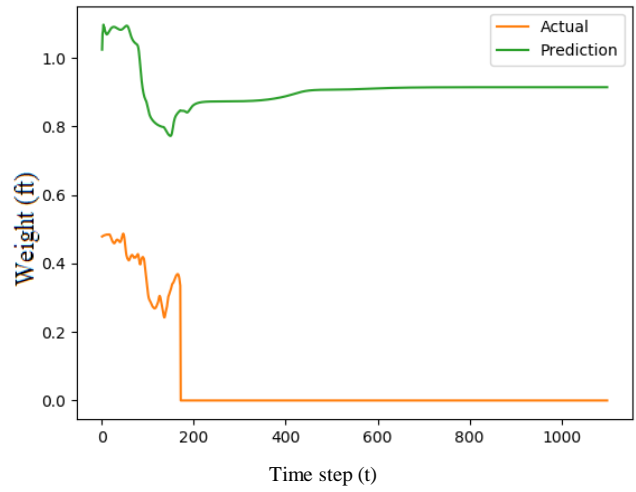
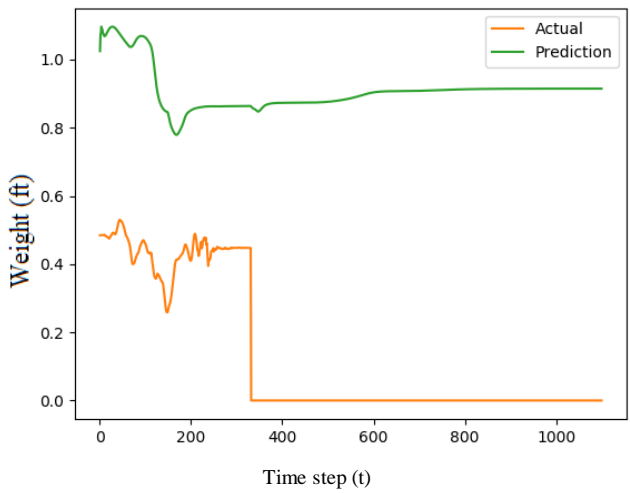
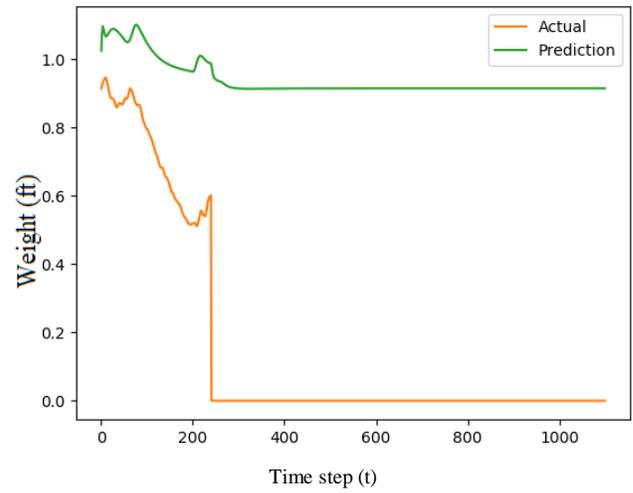
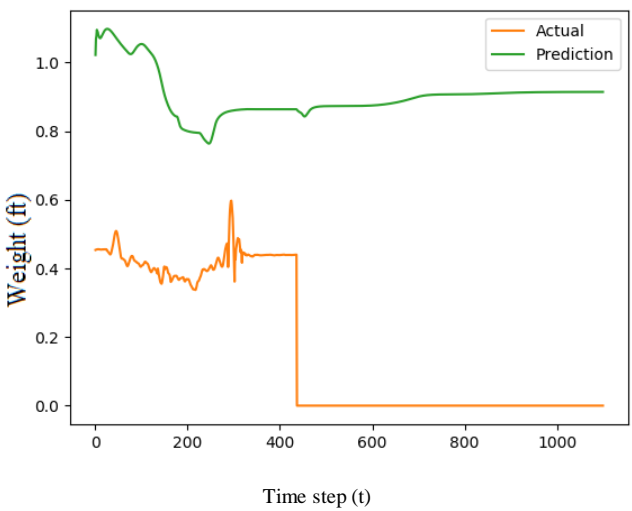
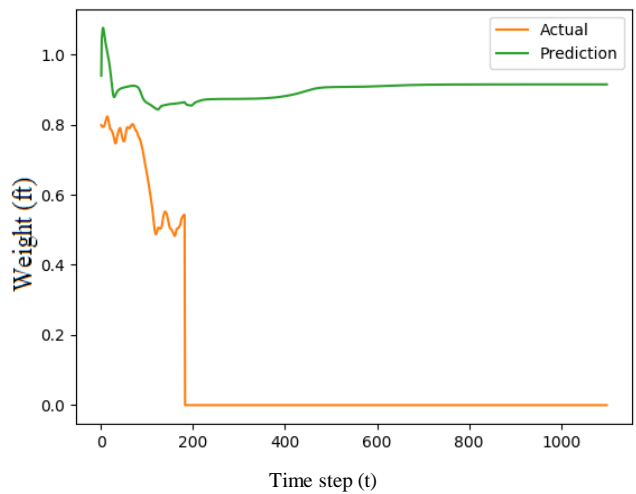

Figure 7: Actual and predicted values of weight ($f_t$) using the provided test set.

Upon analyzing these results, it can be said that the model works fine on the unseen data which does not have much difference in properties like dimension of cup (height and diameter) from test data. On the other hand, in case of generalized data, the model is able to give the same pattern as the target, but there is a considerable difference in the value of $f_t$. Hence, we can say that the model does not perform well for generalization. Reason of such behavior of the model is that the test data has much difference in dimension of cup as compared to the training data. The loss obtained while testing on this data set is 0.7713.

*B. Observation*

In all the graphs plotted during test 1 and test 2, it can be observed that the predicted value of weight ($f_t$)has a similar pattern. Right at the beginning of test sequence, the value of $f_t$ takes a sharp upward spike and then start following the desired pattern i.e. starts changing gradually with time. This is because of sharp decrease in value of the rotation angle $\theta_t$ which directly impacts the value of $f_t$. Upon looking at the data set, it can be observed that for all the sequences, $\theta_t$ decreases sharply for few time steps and then becomes negative. After this point, the sharp change in rotation angle stops and it start changing with small values in both positive and negative directions. This impacts the value of $f_t$ as well and ft start changing with smaller values.

## VI. FUTURE WORK

There are few shortcomings in the experiments which impacted the results. Main problem is that the model does not generalize well. One way to improve generalization capability of the model is to train it on more amount of data with more variation in properties for example dimension of pouring and receiving cup or liquid density. Model may have problem to generalize because of over-fitting. Limiting the capacity of network is one good way to avoid over-fitting. Further experiments can be done with limiting the capacity of by reducing number of hidden layers or reducing number of units per layer. Early stopping the training can be tried to prevent over-fitting. This approach stops the training before the model overfits [28].

## VII. DISCUSSION

In this report, we designed a recurrent neural network which estimates a response to a sequence of manipulation actions. The experiments were done on pouring motion. This approach determines the change in amount of water in the pouring cup with the change in the angle by which the cup is rotated. The model works good with the data similar to the training data, but it does not give good results with more generalized data.

We discussed few ways which can improve the results for generalization. In deep learning, higher is the number of data, higher is the ability of trained model to work well [29]. The model can be improved by using a larger dataset with enhanced variety of instances.

Dynamic trajectory generation and optimization is one of the most important topics in robotics which is being explored widely. There are a plenty of researches going on for generating dynamic response of motion sequences. The results of this review can be helpful for manufacturers who are working in robotics and researchers.